\documentclass[10pt,twocolumn,letterpaper]{article}

\usepackage{cvpr}
\usepackage{times}
\usepackage{epsfig}
\usepackage{graphicx}
\usepackage{amsmath}
\usepackage{amssymb}
\usepackage{subfig}
\usepackage{multirow}
\usepackage{algorithm}
\usepackage{algorithmic}
\usepackage[titletoc]{appendix}

\usepackage[breaklinks=true,bookmarks=false]{hyperref}

\cvprfinalcopy 

\def\cvprPaperID{****} 

\begin{document}

\title{BEVDet4D: Exploit Temporal Cues in Multi-camera 3D Object Detection}

\author{Junjie Huang\thanks{Corresponding author.} \quad Guan Huang \\
       PhiGent Robotics\\
       {\tt\small junjie.huang@ieee.org, guan.huang@phigent.ai}
}

\maketitle

\begin{abstract}
Single frame data contains finite information which limits the performance of the existing vision-based multi-camera 3D object detection paradigms. For fundamentally pushing the performance boundary in this area, a novel paradigm dubbed \textbf{BEVDet4D} is proposed to lift the scalable BEVDet paradigm from the spatial-only 3D working space into the spatial-temporal 4D working space. We upgrade the naive BEVDet framework with a few modifications just for fusing the feature from the previous frame with the corresponding one in the current frame. In this way, with negligible additional computing budget, we enable BEVDet4D to access the temporal cues by querying and comparing the two candidate features. Beyond this, we simplify the task of velocity prediction by degenerating it into the positional offset prediction in the two adjacent features. As a result, BEVDet4D with robust generalization performance reduces the velocity error by up to -62.9\%. This makes the vision-based methods, for the first time, become comparable with those relied on LiDAR or radar in this aspect. On challenge benchmark nuScenes, we report a new record of 54.5\% NDS with the high-performance configuration dubbed BEVDet4D-Base. At the same inference speed, this notably surpasses the previous leading method BEVDet-Base by +7.3\% NDS. The source code is publicly available for further research\footnote {https://github.com/HuangJunJie2017/BEVDet}.
\end{abstract}

\section{Introduction}
Recently, autonomous driving draws great attention in both the research and the industry community. The vision-based perception tasks in this scene include 3D object detection, BEV semantic segmentation, motion prediction, and so on. Most of them can be partly solved in the spatial-only 3D working space with a single frame of data. However, with respect to the time-relevant targets like velocity, current vision-based paradigms with merely a single frame of data perform far poorer than those with sensors like LiDAR or radar. For example, the velocity error of the recently leading method BEVDet \cite{BEVDet} in the vision-based 3D object detection is 3 times that of the LiDAR-based method CenterPoint \cite{CenterPoint3D} and 2 times that of the radar-based method CenterFusion \cite{Centerfusion}. To close this gap, we propose a novel paradigm dubbed BEVDet4D in this paper and pioneer the exploitation of vision-based autonomous driving in the spatial-temporal 4D space.

\begin{figure}[t]
		\centering
		\includegraphics[width=0.78\linewidth]{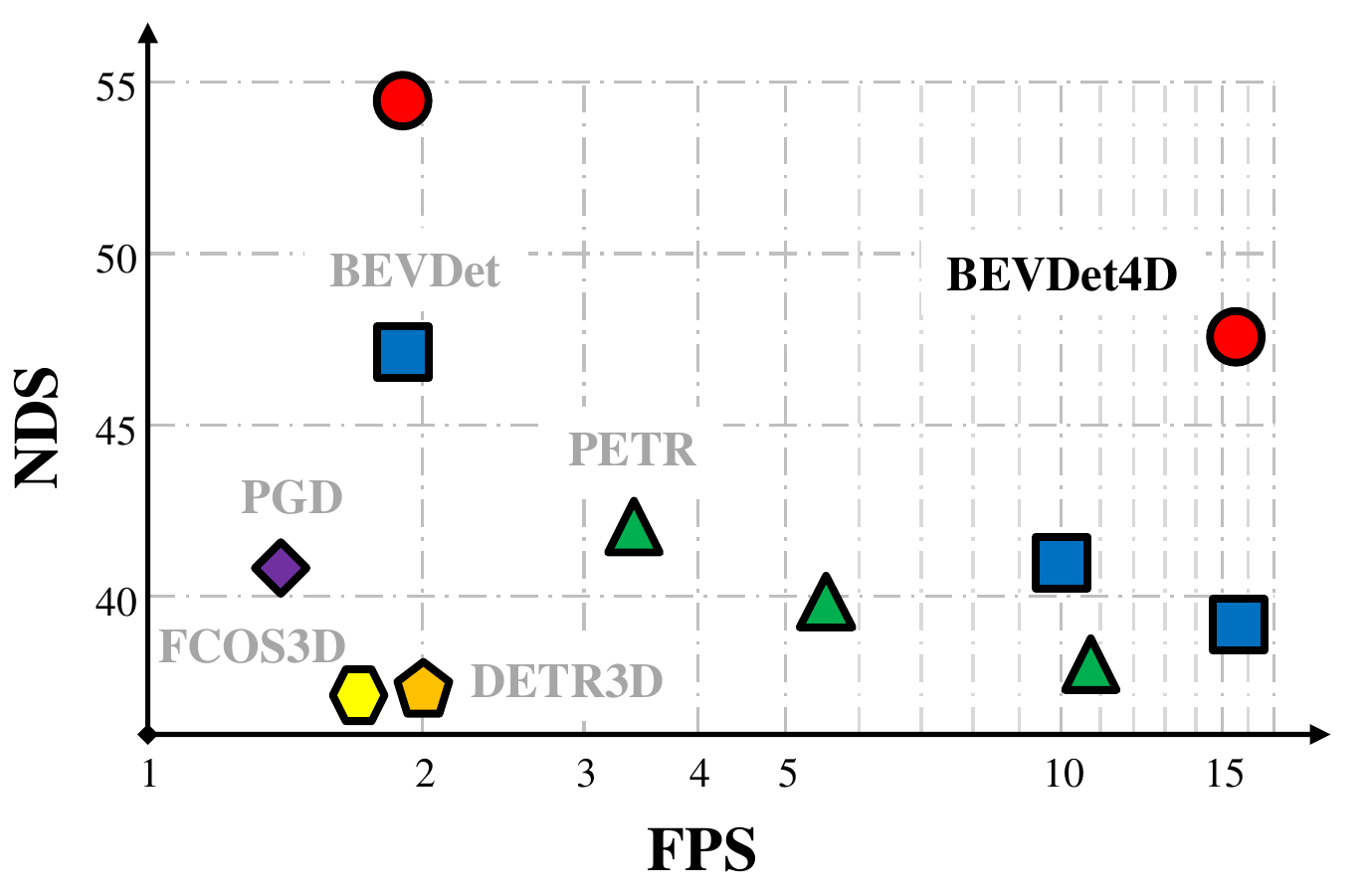}
		\caption{The inference speed and performance of different paradigms on the nuScenes \texttt{val} set.}
		\label{fig:map-flops}
\end{figure}

\begin{figure*}[t]
		\centering
		\includegraphics[width=0.73\linewidth]{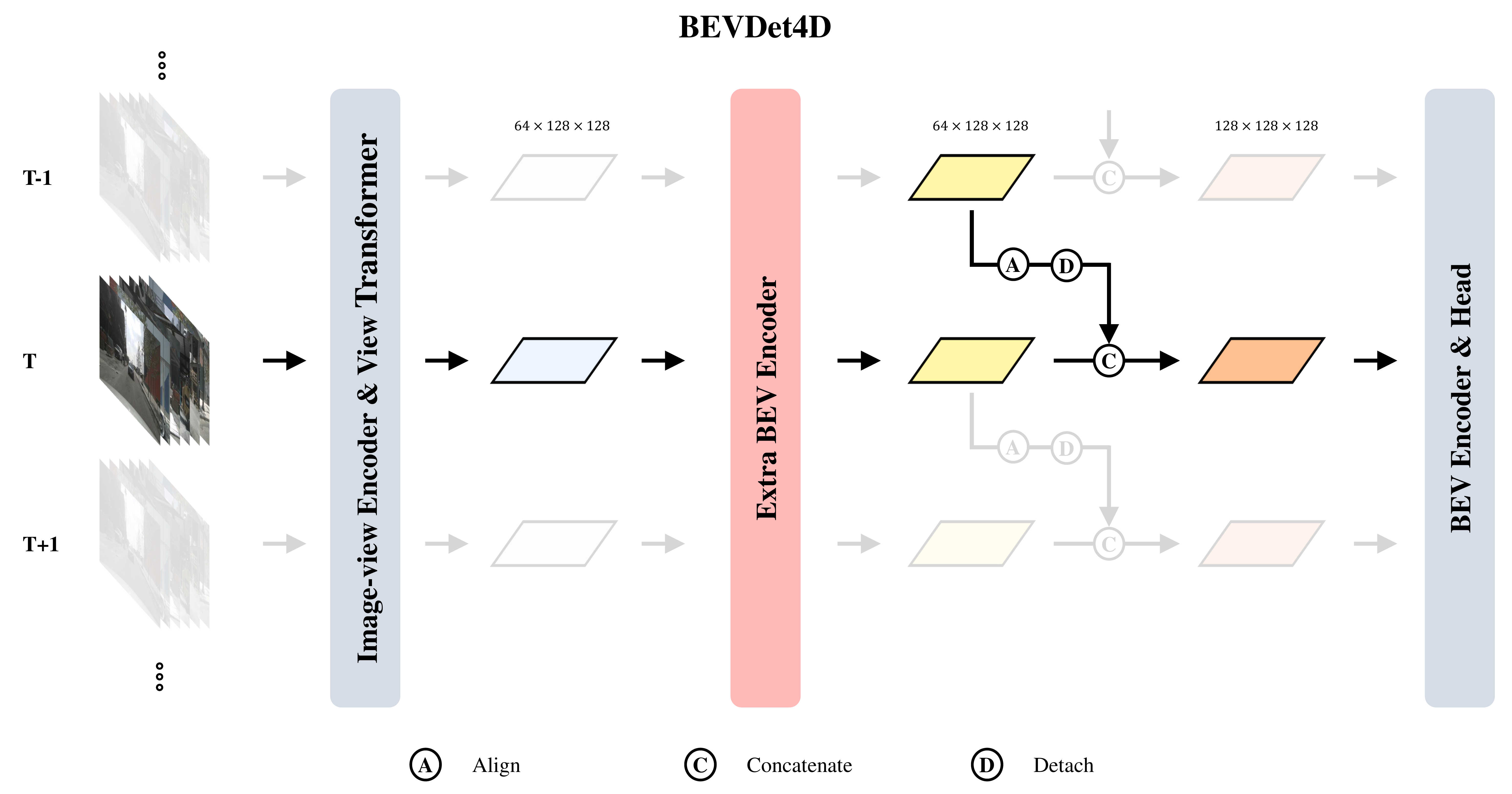}
		\caption{The framework of the proposed BEVDet4D paradigm. BEVDet4D retains the intermediate BEV feature of the previous frame and concatenates it with the ones generated by the current frame. Before that spatial alignment in the flat plane is conducted to partially simplify the velocity prediction task.}
		\label{fig:pipeline}
\end{figure*}

As illustrated in Fig.~\ref{fig:pipeline}, BEVDet4D makes the first attempt at accessing the rich information in the temporal domain. It simply extends the naive BEVDet \cite{BEVDet} by retaining the intermediate BEV features in the previous frames. Then it fuses the retained feature with the corresponding one in the current frame just by a spatial alignment operation and a concatenation operation. Other than that, we kept most other details of the framework unchanged. In this way, we place just a negligible extra computational budget on the inference process while enabling the paradigm to access the temporal cues by querying and comparing the two candidate features. Though simple in constructing the framework of BEVDet4D, it is nontrivial to build its robust performance. The spatial alignment operation and the learning targets should be carefully designed to cooperate with the elegant framework so that the velocity prediction task can be simplified and superior generalization performance can be achieved with BEVDet4D.

We conduct comprehensive experiments on the challenge benchmark nuScenes \cite{NS} to verify the feasibility of BEVDet4D and study its characteristics. Fig.~\ref{fig:map-flops} illustrates the trade-off between inference speed and performance of different paradigms. Without bells and whistles, the BEVDet4D-Tiny configuration reduces the velocity error by 62.9\% from 0.909 mAVE to 0.337 mAVE. Besides, the proposed paradigm also has significant improvement in the other indicators like detection score (+2.6\% mAP), orientation error (-12.0\% mAOE), and attribute error (-25.1\% mAAE). As a result, BEVDet4D-Tiny exceeds the baseline by +8.4\% on the composite indicator NDS. The high-performance configuration dubbed BEVDet4D-Base scores high as 42.1\% mAP and 54.5\% NDS, which has surpassed all published results in vision-based 3D object detection \cite{FCOS3D,PGD,DETR3D,PETR,Graph-DETR3D,BEVFormer,BEVDet}. Last but not least, BEVDet4D achieves the aforementioned superiority just at a negligible cost in inference latency, which is meaningful in the scenario of autonomous driving.

\section{Related Works}

\subsection{Vision-based 3D object detection}
Vision-based 3D object detection is a promising perception task in autonomous driving. In the last few years, fueled by the KITTI \cite{KITTI} benchmark monocular 3D object detection has witness a rapid development \cite{lu2021geometry,liu2021autoshape,zhang2021objects,zou2021devil,zhou2021monocular,reading2021categorical,wang2021progressive,wang2021depth,kumar2021groomed}. However, the limited data and the single view disable it in developing more complicated tasks. Recently, some large-scale benchmarks \cite{NS, Waymo} have been proposed with sufficient data and surrounding views, offering new perspectives toward the paradigm development in the field of 3D object detection. Based on these benchmarks, some multi-camera 3D object detection paradigms have been developed with competitive performance. For example, inspired by the success of FCOS \cite{FCOS} in 2D detection, FCOS3D \cite{FCOS3D} treats the 3D object detection problem as a 2D object detection problem and conducts perception just in image view. Benefitting from the strong spatial correlation of the targets' attribute with the image appearance, it works well in predicting this but is relatively poor in perceiving the targets' translation, velocity, and orientation. PGD \cite{PGD} further develops the FCOS3D paradigm by searching and resolving the outstanding shortcoming (\textit{i.e.} the prediction of the targets' depth). This offers a remarkable accuracy improvement on the baseline but at the cost of more computational budget and additional inference latency. Following DETR \cite{DETR}, DETR3D \cite{DETR3D} proposes to detect 3D objects in an attention pattern, which has similar accuracy as FCOS3D. Although DETR3D requires just half the computational budget, the complex calculation pipeline slows down its inference speed to the same level as FCOS3D. PETR \cite{PETR} further develops the performance of this paradigm by introducing the 3D coordinate generation and position encoding. Besides, they also exploit the strong data augmentation strategies just as BEVDet \cite{BEVDet}. Another concurrent work dubbed Graph-DETR3D \cite{Graph-DETR3D} also extends the DETR3D from two expects. Analogous to the second stage in CenterPoint \cite{CenterPoint3D}, Graph-DETR3D samples multiple points in the 3D space instead of a single point when generating the features of the object queries. Another modification is making the multi-scale training become feasible for DETR3D paradigm by dynamically adjusting the depth target according to the scaling factor. As a novel paradigm, BEVDet \cite{BEVDet} makes the first attempt at applying a strong data augmentation strategy in vision-based 3D object detection. As BEVDet explicitly encodes features in the BEV space, it is scalable in multiple aspects including multi-tasks learning, multi-sensors fusion, and temporal fusion. BEVDet4D is the temporal extension of BEVDet \cite{BEVDet}.

So far, few works have exploited the temporal cues in vision-based 3D object detection. Thus, the existing paradigms \cite{FCOS3D, PGD, DETR3D, PETR, BEVDet} perform relatively poorly in predicting the time-relevant targets like velocity than the LiDAR-based \cite{CenterPoint3D} or radar-based \cite{Centerfusion} methods. To the best of our knowledge, \cite{brazil2020kinematic} is the only one pioneer in this perspective. However, they predict the results based on a single frame and exploit the 3D Kalman filter to update the results for the temporal consistency of results between image sequences. The temporal cues are exploited in the post-processing phase instead of the end-to-end learning framework. Differently, we make the first attempt in exploiting the temporal cues in the end-to-end learning framework BEVDet4D, which is elegant, powerful, and still scalable. BEVFormer \cite{BEVFormer} is a concurrent work of BEVDet4D. Analogous to those \cite{chen2020memory, deng2019object, deng2019relation} in the VID literature, they mainly focus on the feature fusion in the spatial-temporal 4D working space with the attention mechanism \cite{Transformer}. The comparable velocity precision of BEVFormer is achieved by fusing features from multiple adjacent frames (\textit{i.e.} 4 frames in total), which is analogous to most LiDAR-based methods \cite{NS, CenterPoint3D} with points from multiple sweeps. This is fundamentally different from the proposed BEVDet4D, which uses merely two adjacent frames and achieved a higher velocity precision in a more elegant pattern.

\subsection{Object Detection in Video}
Video object detection mainly fueled by the ImageNet VID dataset \cite{ImageNetVID} is analogous to the well-known tasks of common object detection \cite{COCO} which performs and evaluates the object detection task in the image-view space. The difference is that detecting objects in video can access the temporal cues for improving detection accuracy. The methods in this area access the temporal cues mainly according to two kinds of mediums: the predicting results or the intermediate features. The former \cite{TAFVOD} is analogous to \cite{brazil2020kinematic} in vision-based 3D object detection, who optimizes the prediction results in a tracking pattern. The latter reutilizes the features from the previous frame based on some special architectures like LSTM \cite{LSTM} for feature distillation \cite{liu2019looking,liu2018mobile,lu2017online}, attention mechanism \cite{Transformer} for feature querying \cite{chen2020memory, deng2019object, deng2019relation}, and optical flow \cite{Flownet} for feature alignment \cite{zhu2017flow, zhu2018towards}. Specific for the scene of autonomous driving, BEVDet4D is analogous to the flow-based methods in mechanism but accesses the spatial correlation according to the ego-motion and conducts feature aggregation in the 3D space. Besides, BEVDet4D mainly focuses on the prediction of the velocity targets which is not in the scope of the common video object detection literature.

\begin{figure*}[t]
		\centering
		\includegraphics[width=0.8\linewidth]{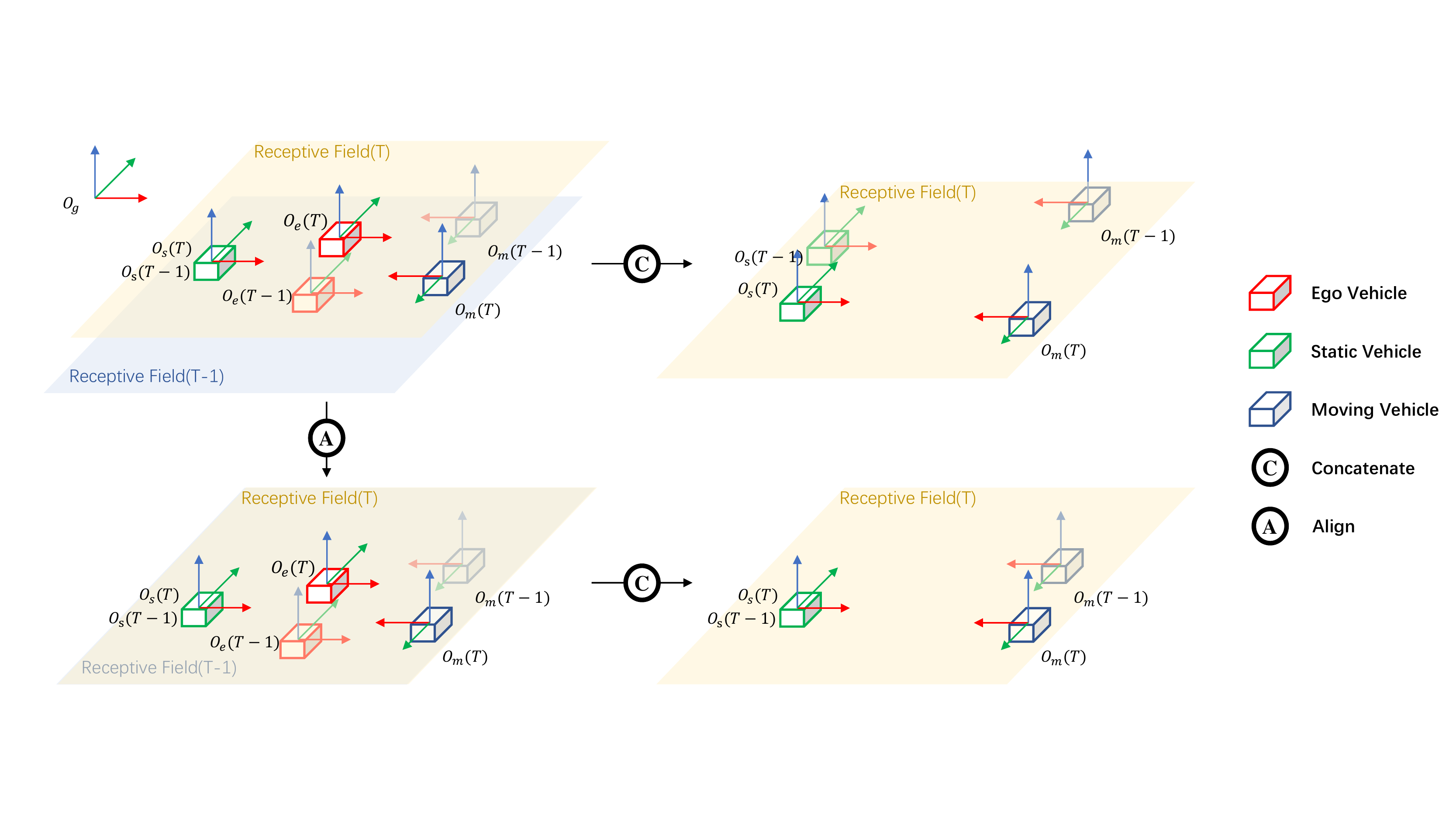}
		\caption{Illustrating the effect of the feature alignment operation. Without the alignment operation (\textit{i.e.} the first row), the following modules are required to study a more complicated distribution of the object motion, which is relevant to the ego-motion. By applying alignment operation in the second row, the learning targets can be simplified.}
		\label{fig:align}
\end{figure*}

\section{Methodology}

\subsection{Network Structure}
As illustrated in Fig.~\ref{fig:pipeline}, the overall framework of BEVDet4D is built upon the BEVDet \cite{BEVDet} baseline which is consists of four kinds of modules: an image-view encoder, a view transformer, a BEV encoder, and a task-specific head. All implementation details of these modules are kept unchanged. To exploit the temporal cues, BEVDet4D extends the baseline by retaining the BEV features generated by the view transformer in the previous frame. Then the retained feature is merged with the one in the current frame. Before that, an alignment operation is conducted to simplify the learning targets which will be detailed in the following subsection. We apply a simple concatenation operation to merge the features for verifying the BEVDet4D paradigm. More complicated fusing strategies have not been exploited in this paper.

Besides, the feature generated by the view transformer is sparse, which is too coarse for the subsequential modules to exploit the temporal cues. Therefore, an extra BEV encoder is applied to adjust the candidate features before the temporal fusion. In practice, the extra BEV encoder consists of two naive residual units \cite{ResNet}, whose channel number is set the same as the input feature.

\subsection{Simplify the Velocity Learning Task}
\paragraph{Symbol Definition} Following nuScense \cite{NS}, we denote the global coordinate system as $O_g-XYZ$, the ego coordinate system as $O_{e(T)}-XYZ$, and the targets coordinate system as $O_{t(T)}-XYZ$. As illustrated in Fig.~\ref{fig:align}, we construct a virtual scene with a moving ego vehicle and two target vehicles. One of the targets is static (\textit{i.e.}, $O_{s}-XYZ$ painted green) in the global coordinate system, while the other one is moving (\textit{i.e.}, $O_{m}-XYZ$ painted blue). The objects in two adjacent frames (\textit{i.e.}, frame $T-1$ and frame $T$) are distinguished with different transparentness. The position of the objects is formulated as $\textbf{P}^x(t)$. $x\in\{g,e(T),e(T-1)\}$ denotes the coordinate system where the position is defined in. $t\in\{T,T-1\}$ denotes the time when the position is recorded. We use $\textbf{T}^{dst}_{src}$ to denote the transformation from the source coordinate system into the target coordinate system.

Instead of directly predicting the velocity of the targets, we tend to predict the translation of the targets in the two adjacent frames. In this way, the learning task can be simplified as the time factor is removed and the positional shifting can be measured just according to the difference between the two BEV features. Besides, we tend to learn the position shifting that is irrelevant to the ego-motion. In this way, the learning task can also be simplified as the ego-motion will make the distribution of the targets' positional shifting more complicated.

For example, due to the ego-motion, a static object (\textit{i.e.}, the green box in Fig.~\ref{fig:align}) in the global coordinate system will be changed into a moving object in the ego coordinate system. More specifically, the receptive field of the BEV features is symmetrically defined around the ego. Considering the two features generated by the view transformer in the two adjacent frames, their receptive fields in the global coordinate system are diverse due to the ego-motion. Given a static object, its position in the global coordinate system is denoted as $\textbf{P}^g_{s}(T)$ and $\textbf{P}^g_{s}(T-1)$ in the two adjacent frames. The positional shifting in the two features should be formulated as:
\begin{equation}
\label{eq:target}
    \begin{split}
          &\textbf{P}^{e(T)}_{s}(T)-\textbf{P}^{e(T-1)}_{s}(T-1) \\
        = &\textbf{T}^{e(T)}_g\textbf{P}^g_{s}(T)-\textbf{T}^{e(T-1)}_g\textbf{P}^g_{s}(T-1)\\
        = &\textbf{T}^{e(T)}_g\textbf{P}^g_{s}(T)-\textbf{T}^{e(T-1)}_{e(T)}\textbf{T}^{e(T)}_g\textbf{P}^g_{s}(T-1)\\
    \end{split}
\end{equation}
According to Eq.~\ref{eq:target}, if we directly concatenate the two features, the learning target (\textit{i.e.}, the positional shifting of the target in the two features) of the following modules is relevant to the ego motion (\textit{i.e.}, $\textbf{T}^{e(T-1)}_{e(T)}$). To avoid this, we shift the target in the adjacent frame by $\textbf{T}^{e(T)}_{e(T-1)}$ to remove the fraction of ego-motion.
\begin{equation}
\label{eq:target_align}
    \begin{split}
          &\textbf{P}^{e(T)}_{s}(T)-\textbf{T}^{e(T)}_{e(T-1)}\textbf{P}^{e(T-1)}_{s}(T-1) \\
        = &\textbf{T}^{e(T)}_g\textbf{P}^g_{s}(T)-\textbf{T}^{e(T)}_{e(T-1)}\textbf{T}^{e(T-1)}_{e(T)}\textbf{T}^{e(T)}_g\textbf{P}^g_{s}(T-1)\\
        = &\textbf{T}^{e(T)}_g\textbf{P}^g_{s}(T)-\textbf{T}^{e(T)}_g\textbf{P}^g_{s}(T-1)\\
        = &\textbf{P}^{e(T)}_{s}(T)-\textbf{P}^{e(T)}_{s}(T-1)\\
    \end{split}
\end{equation}
According to Eq.~\ref{eq:target_align}, the learning target is set as the object's motion in the current frame's ego coordinate system, which is irrelevant to the ego-motion.

\begin{table*}[t]
  \centering
  \caption{Comparison of different paradigms on the nuScenes \texttt{val} set. $\dag$ initialized from a FCOS3D backbone. $\S$ with test-time augmentation. $\#$ with model ensemble.}
  	\resizebox{1.0\linewidth}{!}{
    \begin{tabular}{l|crrc|cccccc|c|r}
    \hline

    \hline
    Methods                     &Image Size         & \#param.  &GFLOPs     & Modality & \textbf{mAP}$\uparrow$ & mATE$\downarrow$  & mASE$\downarrow$   & mAOE$\downarrow$  & mAVE$\downarrow$  &  mAAE$\downarrow$ & \textbf{NDS}$\uparrow$   &FPS\\
    \hline
    CenterFusion \cite{Centerfusion} &800$\times$450&20.4M      &-         & Camera \& Radar& 0.332   & 0.649             & 0.263            & 0.535             & 0.540             & 0.142            & 0.453           &- \\
    VoxelNet \cite{CenterPoint3D}&-                 & 8.8M      &-         & LiDAR    &0.563          &0.292              &0.253             &0.316             &0.287             &0.191            &0.648            &14.1\\
    PointPillar \cite{CenterPoint3D}&-              & 6.0M      &-         & LiDAR    &0.487          &0.315              &0.260               &0.368             &0.323             &0.203            &0.597            &17.9\\
    \hline
    CenterNet\cite{CenterNet}  &-                  &-          &-          & Camera   & 0.306         & 0.716             & 0.264              & 0.609             & 1.426             & 0.658             & 0.328           &-\\
    FCOS3D \cite{FCOS3D}       &1600$\times$900    &52.5M      &2,008.2    & Camera   & 0.295         & 0.806             & 0.268              & 0.511             & 1.315             & \textbf{0.170}    & 0.372           &1.7\\
    DETR3D \cite{DETR3D}       &1600$\times$900    &51.3M      &1,016.8    & Camera   & 0.303         & 0.860             & 0.278              & 0.437             & 0.967             & 0.235             & 0.374           &2.0\\
    PGD \cite{PGD}             &1600$\times$900    &53.6M      &2,223.0    & Camera   & 0.335         & 0.732             & 0.263              & 0.423             & 1.285             & 0.172             & 0.409           &1.4\\
    PETR-R50  \cite{PETR}      &1056$\times$384    &-         &-           & Camera   & 0.313         & 0.768             & 0.278              & 0.564             & 0.923             & 0.225             & 0.381           &10.7 \\
    PETR-R101 \cite{PETR}      &1408$\times$512    &-         &-           & Camera   & 0.357         & 0.710             & 0.270              & 0.490             & 0.885             & 0.224             & 0.421           &3.4 \\
    PETR-Tiny \cite{PETR}      &1408$\times$512    &-         &-           & Camera   & 0.361         & 0.732             & 0.273              & 0.497             & 0.808             & 0.185             & 0.431           &- \\
    BEVDet-Tiny\cite{BEVDet}   &704$\times$256     &52.6M      &215.3      & Camera   & 0.312         & 0.691             & 0.272              & 0.523             & 0.909             & 0.247             & 0.392           &\textbf{15.6}\\
    BEVDet-Base\cite{BEVDet}   &1600$\times$640    &126.6M     &2,962.6    & Camera   & 0.393         & 0.608             & 0.259              & 0.366             & 0.822             & 0.191             & 0.472           &1.9\\
    BEVFormer \cite{BEVFormer} &1600$\times$900   &-          &-          & Camera   & 0.416         & 0.673             & 0.274              & 0.372             & 0.394             & 0.198             & 0.517           &1.7\\
    \textbf{BEVDet4D-Tiny}     &704$\times$256     &53.6M      &222.0      & Camera   & 0.338         & 0.672             & 0.274              & 0.460             & 0.337             & 0.185             & 0.476           &15.5\\
    \textbf{BEVDet4D-Base}     &1600$\times$640    &127.6M     &2,989.2    & Camera   & \textbf{0.421}& \textbf{0.579}    & \textbf{0.258}     & \textbf{0.329}    & \textbf{0.301}    & 0.191             & \textbf{0.545}  &1.9\\

    \hline
    FCOS3D$\dag\S\#$ \cite{FCOS3D}&1600$\times$900  &-          &-          & Camera   & 0.343         & 0.725             & 0.263              & 0.422             & 1.292             & \textbf{0.153}    & 0.415           &-\\
    DETR3D$\dag$ \cite{DETR3D}  &1600$\times$900    &51.3M      &-          & Camera   & 0.349         & 0.716             & 0.268              & 0.379             & 0.842             & 0.200             & 0.434           &-\\
    PGD$\dag\S$ \cite{PGD}      &1600$\times$900    &53.6M      &-          & Camera   & 0.369         & 0.683             & 0.260              & 0.439             & 1.268             & 0.185             & 0.428           &-\\
    PETR-R101$\dag$ \cite{PETR}&1600$\times$900     &-         &-           & Camera   & 0.370         & 0.711             & 0.267              & 0.383             & 0.865             & 0.201             & 0.442           &-\\
    BEVDet-Base$\S$\cite{BEVDet}&1600$\times$640    &126.6M     &-          & Camera   & 0.397         & 0.595             & 0.257              & 0.355             & 0.818             & 0.188             & 0.477           &-\\
    \textbf{BEVDet4D-Base}$\S$ &1600$\times$640    &126.6M     &-           & Camera   & \textbf{0.426}& \textbf{0.560}    & \textbf{0.254}     & \textbf{0.317}    & \textbf{0.289}    & 0.186             & \textbf{0.552}  &-\\
    \hline

    \hline
    \end{tabular}%
    }
  \label{tab:nus-val}%
\end{table*}%

In practice, the alignment operation in Eq.~\ref{eq:target_align} is achieved by feature alignment. Given the candidate features of the previous frame $\mathcal{F}(T-1,\textbf{P}^{e(T-1)})$ and the current frame $\mathcal{F}(T,\textbf{P}^{e(T)})$, the aligned feature can be obtained by:
\begin{equation}
\label{eq:feature_align}
    \mathcal{F}'(T-1,\textbf{P}^{e(T)}) = \mathcal{F}(T-1,\textbf{T}^{e(T-1)}_{e(T)}\textbf{P}^{e(T)})
\end{equation}
Alone with Eq.~\ref{eq:feature_align}, bilinear interpolation is applied as $\textbf{T}^{e(T-1)}_{e(T)}\textbf{P}^{e(T)}$ may not be a valid position in the sparse feature of $\mathcal{F}(T-1,\textbf{P}^{e(T-1)})$. The interpolation is a sub-optimal method that will lead to precision degeneration. The magnitude of the precision degeneration is negatively correlated with the resolution of the BEV features. A more precise method is to adjust the coordinates of the point cloud generated by the lifting operation in the view transformer \cite{LSS}. However, it is deprecated in this paper as it will destroy the precondition of the acceleration method proposed in the naive BEVDet \cite{BEVDet}.  The magnitude of the precision degeneration will be quantitatively estimated in the ablation study Section.~\ref{sec:precisondegeneration}.

\section{Experiment}
\subsection{Experimental Settings}
\paragraph{Dataset} We conduct comprehensive experiments on a large-scale dataset, nuScenes \cite{NS}. nuScenes dataset includes 1000 scenes with images from 6 cameras with surrounding views, points from 5 Radars and 1 LiDAR. It is the up-to-date popular benchmark for 3D object detection \cite{FCOS3D,DETR3D,PGD,DD3D} and BEV semantic segmentation \cite{PON,LSS,VPN,PYVA}. The scenes are officially split into 700/150/150 scenes for training/validation/testing. There are up to 1.4M annotated 3D bounding boxes for 10 classes: car, truck, bus, trailer, construction vehicle, pedestrian, motorcycle, bicycle, barrier, and traffic cone. Following CenterPoint \cite{CenterPoint3D}, we define the region of interest (ROI) within 51.2 meters in the ground plane with a resolution of 0.8 meters by default.

\paragraph{Evaluation Metrics} For 3D object detection, we report the official predefined metrics: mean Average Precision (mAP), Average Translation Error (ATE), Average Scale Error (ASE), Average Orientation Error (AOE), Average Velocity Error (AVE), Average Attribute Error (AAE), and NuScenes Detection Score (NDS). The mAP is analogous to that in 2D object detection \cite{COCO} for measuring the precision and recall, but defined based on the match by 2D center distance on the ground plane instead of the Intersection over Union (IOU) \cite{NS}. NDS is the composite of the other indicators for comprehensively judging the detection capacity. The remaining metrics are designed for calculating the positive results' precision on the corresponding aspects (\emph{e.g.}, translation, scale, orientation, velocity, and attribute).

\begin{table*}[t]
  \centering
  \caption{Comparison with the state-of-the-art methods on the nuScenes \texttt{test} set. $\dag$ pre-train on DDAD \cite{DDAD}.}
  	\resizebox{0.85\linewidth}{!}{
    \begin{tabular}{l|c|ccccccc}
    \hline

    \hline
    Methods                     & Modality & mAP$\uparrow$ & mATE$\downarrow$   & mASE$\downarrow$  & mAOE$\downarrow$  & mAVE$\downarrow$  &  mAAE$\downarrow$ & NDS$\uparrow$ \\
    \hline
    PointPillars(Light) \cite{PointPillar}& LiDAR& 0.305    & 0.517              & 0.290             & 0.500             & 0.316             & 0.368             & 0.453          \\
    CenterFusion \cite{Centerfusion}& Camera \& Radar& 0.326& 0.631              & 0.261             & 0.516             & 0.614             & 0.115             & 0.449          \\
    CenterPoint \cite{CenterPoint3D}& Camera \& LiDAR \& Radar& 0.671 & 0.249    & 0.236             & 0.350             & 0.250             & 0.136             & 0.714          \\
    \hline
    MonoDIS \cite{MonoDIS}       & Camera   & 0.304         & 0.738              & 0.263             & 0.546             & 1.553             & 0.134             & 0.384          \\
    CenterNet \cite{CenterNet}   & Camera   & 0.338         & 0.658              & 0.255             & 0.629             & 1.629             & 0.142             & 0.400          \\
    FCOS3D   \cite{FCOS3D}       & Camera   & 0.358         & 0.690              & 0.249             & 0.452             & 1.434              & 0.124             & 0.428          \\
    PGD      \cite{PGD}          & Camera   & 0.386         & 0.626              & 0.245             & 0.451             & 1.509              & 0.127             & 0.448          \\
    PETR \cite{PETR}             & Camera   & 0.434         & 0.641              & 0.248             & 0.437             & 0.894             & 0.143             & 0.481          \\
    BEVDet \cite{BEVDet}         & Camera   & 0.422         & 0.529              & \textbf{0.236}    & 0.395             & 0.979             & 0.152             & 0.482          \\
    BEVFormer \cite{BEVFormer}   & Camera   & 0.445         & 0.631              & 0.257             & 0.405             & 0.435             & 0.143             & 0.535          \\
    \textbf{BEVDet4D}            & Camera   & \textbf{0.451}& \textbf{0.511}     & 0.241             & \textbf{0.386}    & \textbf{0.301}    & \textbf{0.121}    & \textbf{0.569}\\
    \hline
    DD3D$\dag$ \cite{DD3D}       & Camera   & 0.418         & 0.572              & 0.249             & 0.368             & 1.014             & 0.124             & 0.477          \\
    DETR3D$\dag$ \cite{DETR3D}   & Camera   & 0.386         & 0.626              & 0.245             & 0.394             & 0.845             & 0.133             & 0.479          \\
    Graph-DETR3D$\dag$ \cite{Graph-DETR3D}& Camera   & 0.425         & 0.621     & 0.251             & 0.386             & 0.790             & 0.128             & 0.495          \\
    PETR$\dag$ \cite{PETR}       & Camera   & 0.441         & 0.593              & 0.249             & 0.383             & 0.808             & 0.132             & 0.504          \\
    \hline

    \hline
    \end{tabular}%
    }
  \label{tab:nus-test}%
\end{table*}%

\paragraph{Training Parameters}
Following BEVDet \cite{BEVDet}, models are trained with AdamW \cite{AdamW} optimizer, in which gradient clip is exploited with learning rate 2e-4, a total batch size of 64 on 8 NVIDIA GeForce RTX 3090 GPUs. Sublinear memory cost \cite{SMC} is used for GPU memory management. We apply a cyclic policy \cite{Second}, which linearly increases the learning rate from 2e-4 to 1e-3 in the first 40\% schedule and linearly decreases the learning rate from 1e-3 to 0 in the remainder epochs. By default, the total schedule is terminated within 20 epochs.

\paragraph{Data Processing} We keep all data processing settings the same as BEVDet \cite{BEVDet}. Specifically, we use $W_{in}\times H_{in}$ to denote the width and height of the input image. By default in the training process, the source images with 1600$\times$900 resolution \cite{NS} are processed by random flipping, random scaling with a range of $s\in[W_{in}/1600-0.06, W_{in}/1600+0.11]$, random rotating with a range of $r\in[-5.4^{\circ}, 5.4^{\circ}]$, and finally cropping to a size of $W_{in}\times H_{in}$. The cropping is conducted randomly in the horizon direction but is fixed in the vertical direction (\textit{i.e.}, $(y_1,y_2) =(max(0,s*900 - H_{in}), y_1+H_{in})$, where $y_1$ and $y_2$ are the upper bound and the lower bound of the target region.) In the BEV space, the input feature and 3D object detection targets are augmented by random flipping, random rotating with a range of $[-22.5^{\circ}, 22.5^{\circ}]$, and random scaling with a range of $[0.95, 1.05]$. Following CenterPoint \cite{CenterPoint3D}, all models are trained with CBGS \cite{CBGS}. In testing time, the input images are scaled by a factor of $s=W_{in}/1600+0.04$ and cropped to $W_{in}\times H_{in}$ resolution with a region defined as $(x_1,x_2,y_1,y_2)=(0.5*(s*1600-W_{in}), x_1+W_{in}, s*900-H_{in}, y_1+H_{in})$.

\paragraph{Inference Speed} We conduct all experiments based on MMDetection3D \cite{mmdet3d2020}. The inference speed is the average upon 6019 validation samples \cite{NS}. For monocular paradigms like FCOS3D \cite{FCOS3D} and PGD \cite{PGD}, the inference speeds are divided by a factor of 6 (\textit{i.e.} the number of images in a single sample \cite{NS}), as they take each image as an independent sample. By default, the inference acceleration method proposed in BEVDet \cite{BEVDet} is applied.

\subsection{Benchmark Results}

\subsubsection{nuScenes \texttt{val} set}
We comprehensively compare the proposed BEVDet4D with the baseline method BEVDet \cite{BEVDet} and other paradigms like FCOS3D \cite{FCOS3D}, PGD \cite{PGD}, DETR3D \cite{DETR3D}, PETR \cite{PETR} Graph-DETR3D \cite{Graph-DETR3D} and BEVFormer \cite{BEVFormer}. Their numbers of parameters, computational budget, inference speed, and accuracy on the nuScenes \texttt{val} set are all listed in Tab.~\ref{tab:nus-val}. Some state-of-the-art methods with other sensors are also listed for comparison like LiDAR-based method CenterPoint \cite{CenterPoint3D} and radar-based method CenterFusion \cite{Centerfusion}.

The high-speed version dubbed BEVDet4D-Tiny scores 47.6\% NDS on nuScenes \texttt{val} set, which exceeds the baseline (\textit{i.e.} BEVDet-Tiny \cite{BEVDet} with 39.2\% NDS) by a large margin of +8.4\% NDS. The improvement in the composite indicator NDS mainly derives from the reduction of the orientation error, the velocity error, and the attribute error. Specifically, benefitting from the well-designed BEVDet4D paradigm, the velocity error is significantly decreased by -62.9\% from BEVDet-Tiny 0.909 mAVE to BEVDet4D-Tiny 0.337 mAVE. For the first time, the precision of the velocity prediction in the camera-based methods notably exceeds the CenterFusion \cite{Centerfusion} 0.540 mAVE, who relies on the multi-sensor fusion with camera and radar for high precision in this aspect. Besides, at a similar inference speed, velocity precision of BEVDet4D-Tiny is also comparable with the state-of-the-art LiDAR-based method PointPillar \cite{PointPillar} (\textit{i.e.} 17.9 FPS and 0.323 mAVE) implemented in \cite{CenterPoint3D}. With respect to the orientation prediction, the proposed method also reduces the error in this aspect by -12.0\% from BEVDet-Tiny 0.523 mAOE to BEVDet4D-Tiny 0.460 mAOE. This is because the orientation and velocity of the targets are strong-coupled. Analogously, the attribute error is reduced by -25.1\% from BEVDet-Tiny 0.247 mAAE to BEVDet4D-Tiny 0.185 mAAE.

While upgrading the paradigm to BEVDet4D-Base analogous to BEVDet-Base \cite{BEVDet}, the promotion on the baseline is slightly narrowed to +7.3\% on the composite indicator NDS from BEVDet-Base 47.2\% NDS to BEVDet4D-Base 54.5\% NDS. This surpasses the concurrent work of BEVFormer \cite{BEVFormer} by +2.8\% NDS (\textit{i.e.} 54.5\% NDS \textit{v.s.} 51.7\% NDS), while running faster than it in test time (\textit{i.e.} 1.9 FPS \textit{v.s.} 1.7 FPS). With test time augmentation, we further push the performance boundary to 55.2\% NDS. It is worth noting that, thanks to the few framework adjustments, BEVDet4D achieves the aforementioned performance improvement at the cost of negligible extra inference latency.

\begin{table*}[t]
  \centering
  \caption{Results of ablation study on the nuScenes \texttt{val} set. The align operation includes rotation (R) and translation (T). Extra denotes the extra BEV encoder. Aug. denotes the augmentation in time dimension when selecting the adjacent frame.}
    \resizebox{1.0\linewidth}{!}{
    \begin{tabular}{c|ccccc|cc|ccccc|crc}
    \hline

    \hline
    Methods     &Align  & Target    & Extra     & Weight  & Aug.      & mAP$\uparrow$ & NDS$\uparrow$ & mATE$\downarrow$  & mASE$\downarrow$  & mAOE$\downarrow$  & mAVE$\downarrow$  & mAAE$\downarrow$  &\#param.   &GFLOPs &FPS   \\
   \hline
    BEVDet      &       & Speed     &           & 0.2     &           & 0.312         & 0.392         &0.691              &0.272              &0.523              &0.909              &0.247              &52.5M      &215.3& 7.8  \\
    \hline
    A           &       & Speed     &           & 0.2     &           & 0.296         & 0.376         &0.711              &0.274              &0.501              &1.544              &0.234              &52.6M      &215.9& 7.8 \\
    B           &T      & Speed     &           & 0.2     &           & 0.321         & 0.393         &0.672              &0.272              &0.525              &1.186              &0.215              &52.6M      &215.9& 7.8  \\
    C           &T      & Offset    &           & 0.2     &           & 0.320         & 0.440         &0.697              &0.274              &0.514              &0.479              &0.229              &52.6M      &215.9& 7.8  \\
    D           &T      & Offset    &\checkmark & 0.2     &           & 0.323         & 0.449         &0.680              &0.274              &0.480              &0.479              &0.209              &53.6M      &222.0& 7.7  \\
    E           &T      & Offset    &\checkmark & 1.0     &           & 0.322         & 0.452         &0.685              &0.271              &0.496              &0.435              &0.201              &53.6M      &222.0& 7.7  \\
    F           &R\&T   & Offset    &\checkmark & 1.0     &           & 0.321         & 0.461         &0.681              &0.272              &0.461              &0.376              &0.203              &53.6M      &222.0& 7.7  \\
    G           &R\&T   & Offset    &\checkmark & 1.0     &\checkmark & 0.340         & 0.481         &0.660              &0.270              &0.453              &0.328              &0.176              &53.6M      &222.0& 7.7  \\
    \hline

    \hline
    \end{tabular}%
    }

  \label{tab:ablation}%
\end{table*}%

\subsubsection{nuScenes \texttt{test} set}
For the nuScenes \texttt{test} set, we train the BEVDet4D-Base configuration on the \texttt{train} and \texttt{val} sets. A single model with test time augmentation is adopted. As listed in Tab.~\ref{tab:nus-test}, BEVDet4D ranks first on the nuScenes vision-based 3D object detection leader board with a score of 56.9\% NDS, substantially surpassing the previous leading method BEVDet \cite{BEVDet} by +8.7\% NDS. It also exceeds the concurrent work of BEVFormer \cite{BEVFormer} by +3.4\% NDS and significantly exceeds those relied on additional data for pre-training like DD3D \cite{DD3D}, DETR3D \cite{DETR3D}, and PETR \cite{PETR}. Besides the composite indicator, BEVDet4D has leading performance in most other indicators like mAP, mATE, mAOE, mAVE and mATE. With respect to the ability of generalization, the previous leading method BEVDet \cite{BEVDet} has merely +0.5\% performance growth from \texttt{val} set 47.7\% NDS to \texttt{test} set 48.2\% NDS. However, with the same configuration, the performance boosting of BEVDet4D is +1.7\% NDS from \texttt{val} set 55.2\% NDS to \texttt{test} set 56.9\% NDS. This indicates that exploiting temporal cues in BEVDet4D can also help improve the models' generalization performance.

\subsection{Ablation Studies}

\subsubsection{Road Map of Building BEVDet4D}
In this subsection, we empirically show how the robust performance of BEVDet4D is built. BEVDet-Tiny \cite{BEVDet} without acceleration is adopted as a baseline. In other words, the spatial alignment operation in BEVDet4D is conducted within the view transformer by adjusting the pseudo point cloud \cite{LSS}. The results of different configurations are listed in Tab.~\ref{tab:ablation}. Some key factors are discussed one by one in the following.

Directly concatenate the current frame feature with the previous one in configuration Tab.~\ref{tab:ablation} (A), the overall performance drops from 39.2\% NDS to 37.6\% NDS by -1.6\%. This modification degrades the models' performance, especially on the translation and the velocity aspects. We conjecture that, due to the ego-motion, the positional shift of the same static object between the two candidate features will confuse the following modules' judgment on the object position. With respect to the moving object, it is more complicated for the modules to judge out the velocity target defined in the current frame's ego coordinate system \cite{BEVDet} from the positional shift between the two candidate features which is described in Eq.~\ref{eq:target}. As to this end, the module needs to remove the fraction of ego-motion from this positional shift and consider the time factor.

By conducting translation-only align operation in configuration Tab.~\ref{tab:ablation} (B), we enable the modules to utilize the position-aligned candidate features to make better perceptions of the static targets. Besides, the velocity predicting task is simplified by removing the fraction of the ego-motion. As result, the translation error is reduced by -5.4\% to 0.672, which has surpassed the baseline configuration with a translation error of 0.691. Moreover, the velocity error is also reduced by -23.2\% from configuration Tab.~\ref{tab:ablation} (A) 1.544 mAVE to configuration Tab.~\ref{tab:ablation} (B) 1.186 mAVE. However, this velocity error is still larger than that of the baseline configuration. We conjecture that the distribution of the positional shift is far from that of the velocity due to the inconsistent time duration between the two adjacent frames.

\begin{figure}[t]
		\centering
		\includegraphics[width=0.88\linewidth]{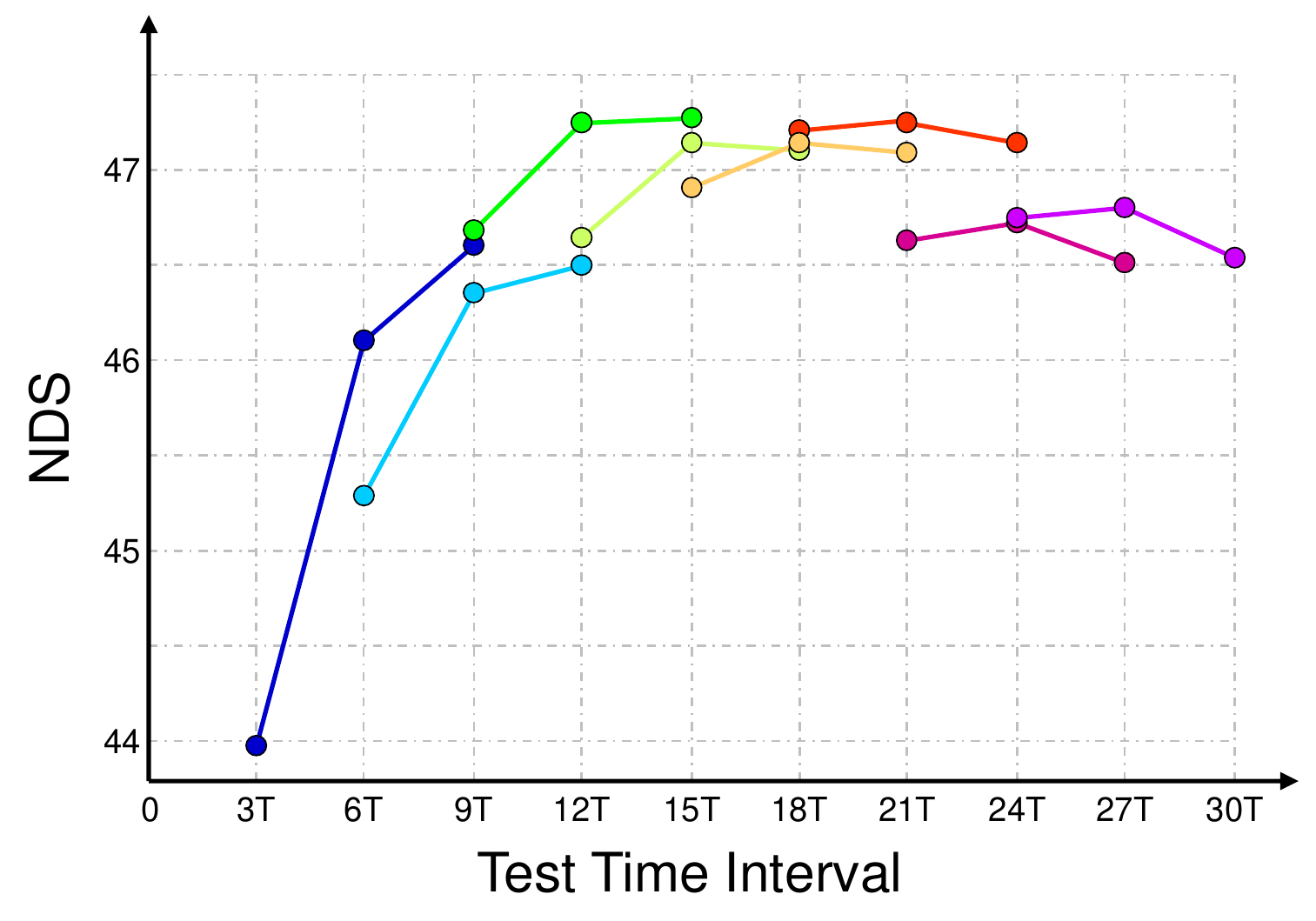}
		\caption{Ablation on the time interval between the current frame and the reference one. Points drawn in the same color are in the same training configuration.}
		\label{fig:interval}
\end{figure}

\begin{table*}[t]
  \centering
  \caption{Ablation study for the precision degeneration of the interpolation operation on the nuScenes \texttt{val} set.}
    \resizebox{1.0\linewidth}{!}{
    \begin{tabular}{c|cc|ccccccc|ccrc}
    \hline

    \hline
    Configuration     &BEV Resolution      & Align     & mAP$\uparrow$ & NDS$\uparrow$ & mATE$\downarrow$  & mASE$\downarrow$  & mAOE$\downarrow$  & mAVE$\downarrow$  & mAAE$\downarrow$  &\#param.   &GFLOPs &FPS \\
    \hline

    \hline
    A           &0.8m$\times$0.8m      & Within    & 0.320         & 0.440         &0.697              &0.274              &0.514              &0.479              &0.229              &52.6M      &215.9& 7.8  \\
    B           &0.8m$\times$0.8m      & After     & 0.313         & 0.438         &0.685              &0.273              &0.529              &0.499              &0.205              &52.6M      &215.9& 15.6 \\
    \hline
    C           &0.4m$\times$0.4m      & Within    & 0.326         & 0.452         &0.651              &0.273              &0.455              &0.516              &0.216              &52.6M      &440.8& 6.8 \\
    D           &0.4m$\times$0.4m      & After     & 0.327         & 0.453         &0.657              &0.271              &0.463              &0.497              &0.219              &52.6M      &440.8& 10.0 \\
    \hline

    \hline
    \end{tabular}%
    }

  \label{tab:precisondegeneration}%
\end{table*}%

\begin{table*}[t]
  \centering
  \caption{Ablation study for the position of temporal fusion in BEVDet4D on the nuScenes \texttt{val} set.}
    \resizebox{0.88\linewidth}{!}{
    \begin{tabular}{c|c|c|cc|ccccc}
    \hline

    \hline
    Methods     &Configuration &Position        & mAP$\uparrow$ & NDS$\uparrow$ & mATE$\downarrow$  & mASE$\downarrow$  & mAOE$\downarrow$  & mAVE$\downarrow$  & mAAE$\downarrow$   \\
    \hline
    BEVDet\cite{BEVDet}& - &           -        & 0.312         & 0.392         & 0.691             &\textbf{0.272}     &0.523              & 0.909             & 0.247             \\
    \hline
    BEVDet4D    &A  &Before Extra BEV Encoder   & 0.320         & 0.442         &0.687              &0.277              &0.519              &0.480              &0.214               \\
    BEVDet4D    &B  &After Extra BEV Encoder    & \textbf{0.323}& \textbf{0.453}&\textbf{0.674}     &\textbf{0.272}     &\textbf{0.503}     & \textbf{0.429}    &\textbf{0.208}      \\
    BEVDet4D    &C  &After BEV Encoder          & 0.311         & 0.394         &0.720              &0.274              &0.536              &0.838              &0.245               \\
    \hline

    \hline
    \end{tabular}%
    }

  \label{tab:pfs}%
\end{table*}%

Further removing the time factor in configuration Tab.~\ref{tab:ablation} (C), we let the module directly predict the targets' positional shift in two candidate features. This modification successfully simplifies the learning targets and makes the trained module more robust on the validation set. The velocity error is thus further reduced by a large margin of -59.6\% to 0.479 mAVE, which is just 52.7\% of the naive BEDVet \cite{BEVDet}.

In configuration Tab.~\ref{tab:ablation} (D), we apply an extra BEV encoder before concatenating the two candidate features. This slightly enlarges the computational budget by 2.8\%. The change of inference speed is negligible. However, this modification offers comprehensive improvement on the baseline (\textit{i.e.}, Tab.~\ref{tab:ablation} (C)). The overall performance is improved by +0.9\% NDS from 44.0\% to 44.9\%. By adjusting the loss weight of velocity prediction in the training process, configuration Tab.~\ref{tab:ablation} (E) reduces the velocity error to 0.435.

By considering the rotation variance of the ego pose in the align operation, configuration Tab.~\ref{tab:ablation} (F) further reduces the velocity error by 13.6\% from 0.435 (\textit{i.e.}, Tab.~\ref{tab:ablation} (E)) to 0.376. This indicates that a precise align operation can help increase the precision of velocity prediction.

To search for the optimal test time interval between the current frame and the reference one, we use the unlabeled camera sweeps with 12Hz instead of the annotated camera frames (2Hz) in configuration Tab.~\ref{tab:ablation} (G). The time interval between two camera sweeps is denoted as $T\approx 0.083s$. We select three different time intervals in each training configuration and judge the adjusting direction by comparing them in test time. In this way, we can avoid the training disturbance in searching for this hyper-parameter. According to Fig.~\ref{fig:interval}, The optimal interval is around 15T which is set as the test time interval by default in this paper. During the training process, we conduct data augmentation by randomly sampling time intervals within $[3T,27T]$. As a result, configuration Tab.~\ref{tab:ablation} (G) further reduces the velocity error by 12.8\% from 0.376 to 0.328.

\subsubsection{Precision Degeneration of the Interpolation}
\label{sec:precisondegeneration}
We use configuration Tab.~\ref{tab:ablation} (C) to exploit the precision degeneration of the interpolation operation. Several ablation configurations are constructed in Tab.~\ref{tab:precisondegeneration} to study the factors like the BEV resolution and the interpolation operation. When a low BEV resolution of 0.8m$\times$0.8m is applied, we observed a slight drop in velocity precision from configuration Tab.~\ref{tab:precisondegeneration} (A) 0.479 mAVE to (B) 0.499 mAVE. This indicates that aligning the feature map after the view transformation with interpolation operation will introduce systematic error. However, the precondition of acceleration in BEVDet \cite{BEVDet} can be maintained in configuration Tab.~\ref{tab:precisondegeneration} (B). Benefitting from the acceleration method, the inference speed can be scaled up to 15.6 FPS, which is twice that of the configuration Tab.~\ref{tab:precisondegeneration} (A).

When a high BEV resolution of 0.4m$\times$0.4m is applied, the performance difference between aligning within the view transformation and aligning after view transformation with interpolation operation is negligible (\textit{i.e.} Tab.~\ref{tab:precisondegeneration} (C) with 45.2 NDS \textit{v.s.} Tab.~\ref{tab:precisondegeneration} (D) with 45.3 NDS). High BEV resolution can help reduce the precision degeneration caused by the interpolation operation. Besides, from the perspective of inference acceleration, conducting aligning operations within the view transformation is deprecated.

\subsubsection{The Position of the Temporal Fusion}
It is not trivial to select the position of the temporal fusion in the BEVDet4D framework. We compare some typical positions in Tab.~\ref{tab:pfs} to study this problem. Among all configurations, conducting temporal fusion after the extra BEV encoder in configuration Tab.~\ref{tab:pfs} (B) is the most applicable one with the lowest velocity error of 0.429 mAVE.  When bringing forward the temporal fusion in configuration Tab.~\ref{tab:pfs} (A), the velocity error is increased by +11.9\% to 0.480 mAVE. This indicates that the BEV feature generated by the view transformer is too coarse to be directly applied. An extra BEV encoder before temporal fusion can help alleviate this problem. When we postpone the temporal fusion to the back of the BEV encoder in configuration Tab.~\ref{tab:pfs} (C), the overall performance degenerates to 39.4\% NDS which is close to the baseline BEVDet \cite{BEVDet} with 39.2\% NDS. More precisely, the feature from the previous frame helps slightly reduce the velocity error from 0.909 mAVE to 0.838 but increases the translation error from 0.691 mATE to 0.720 mATE. This indicates that the BEV encoder plays an important role in effectuating the proposed BEVDet4D paradigm by resisting the positional misleading from the previous frame feature and estimating the velocity according to the difference between the two candidate features.

\section{Conclusion}
We pioneer the exploitation of vision-based autonomous driving in the spatial-temporal 4D space by proposing BEVDet4D to lift the scalable BEVDet \cite{BEVDet} from spatial-only 3D working space into spatial-temporal 4D working space. BEVDet4D retains the elegance of BEVDet while substantially pushing the performance in multi-camera 3D object detection, particularly in the velocity prediction aspect. Future works will focus on the design of framework and paradigm for actively mining the temporal cues.

{\small
\bibliographystyle{ieee_fullname}
\bibliography{egbib}

\begin{thebibliography}{10}\itemsep=-1pt

\bibitem{brazil2020kinematic}
Garrick Brazil, Gerard Pons-Moll, Xiaoming Liu, and Bernt Schiele.
\newblock {Kinematic 3D Object Detection in Monocular Video}.
\newblock In {\em Proceedings of the European Conference on Computer Vision},
  pages 135--152. Springer, 2020.

\bibitem{NS}
Holger Caesar, Varun Bankiti, Alex~H Lang, Sourabh Vora, Venice~Erin Liong,
  Qiang Xu, Anush Krishnan, Yu Pan, Giancarlo Baldan, and Oscar Beijbom.
\newblock {nuScenes: A multimodal dataset for autonomous driving}.
\newblock In {\em Proceedings of the IEEE Conference on Computer Vision and
  Pattern Recognition}, pages 11621--11631, 2020.

\bibitem{DETR}
Nicolas Carion, Francisco Massa, Gabriel Synnaeve, Nicolas Usunier, Alexander
  Kirillov, and Sergey Zagoruyko.
\newblock {End-to-End Object Detection with Transformers}.
\newblock In {\em Proceedings of the European Conference on Computer Vision},
  pages 213--229. Springer, 2020.

\bibitem{SMC}
Tianqi Chen, Bing Xu, Chiyuan Zhang, and Carlos Guestrin.
\newblock {Training Deep Nets with Sublinear Memory Cost}.
\newblock {\em arXiv preprint arXiv:1604.06174}, 2016.

\bibitem{chen2020memory}
Yihong Chen, Yue Cao, Han Hu, and Liwei Wang.
\newblock {Memory Enhanced Global-Local Aggregation for Video Object
  Detection}.
\newblock In {\em Proceedings of the IEEE Conference on Computer Vision and
  Pattern Recognition}, pages 10337--10346, 2020.

\bibitem{Graph-DETR3D}
Zehui Chen, Zhenyu Li, Shiquan Zhang, Liangji Fang, Qinhong Jiang, and Feng
  Zhao.
\newblock {Graph-DETR3D: Rethinking Overlapping Regions for Multi-View 3D
  Object Detection}.
\newblock {\em arXiv preprint arXiv:2204.11582}, 2022.

\bibitem{mmdet3d2020}
MMDetection3D Contributors.
\newblock {MMDetection3D: OpenMMLab} next-generation platform for general {3D}
  object detection.
\newblock \url{https://github.com/open-mmlab/mmdetection3d}, 2020.

\bibitem{deng2019object}
Hanming Deng, Yang Hua, Tao Song, Zongpu Zhang, Zhengui Xue, Ruhui Ma, Neil
  Robertson, and Haibing Guan.
\newblock {Object Guided External Memory Network for Video Object Detection}.
\newblock In {\em Proceedings of the International Conference on Computer
  Vision}, pages 6678--6687, 2019.

\bibitem{deng2019relation}
Jiajun Deng, Yingwei Pan, Ting Yao, Wengang Zhou, Houqiang Li, and Tao Mei.
\newblock {Relation Distillation Networks for Video Object Detection}.
\newblock In {\em Proceedings of the International Conference on Computer
  Vision}, pages 7023--7032, 2019.

\bibitem{Flownet}
Alexey Dosovitskiy, Philipp Fischer, Eddy Ilg, Philip Hausser, Caner Hazirbas,
  Vladimir Golkov, Patrick Van Der~Smagt, Daniel Cremers, and Thomas Brox.
\newblock {FlowNet: Learning Optical Flow with Convolutional Networks}.
\newblock In {\em Proceedings of the International Conference on Computer
  Vision}, pages 2758--2766, 2015.

\bibitem{KITTI}
Andreas Geiger, Philip Lenz, and Raquel Urtasun.
\newblock {Are we ready for Autonomous Driving? The KITTI Vision Benchmark
  Suite}.
\newblock In {\em Proceedings of the IEEE Conference on Computer Vision and
  Pattern Recognition}, 2012.

\bibitem{DDAD}
Vitor Guizilini, Rares Ambrus, Sudeep Pillai, Allan Raventos, and Adrien
  Gaidon.
\newblock {3D Packing for Self-Supervised Monocular Depth Estimation}.
\newblock In {\em Proceedings of the IEEE Conference on Computer Vision and
  Pattern Recognition}, pages 2485--2494, 2020.

\bibitem{ResNet}
Kaiming He, Xiangyu Zhang, Shaoqing Ren, and Jian Sun.
\newblock {Deep Residual Learning for Image Recognition}.
\newblock In {\em Proceedings of the IEEE Conference on Computer Vision and
  Pattern Recognition}, pages 770--778, 2016.

\bibitem{LSTM}
Sepp Hochreiter and J{\"u}rgen Schmidhuber.
\newblock Long short-term memory.
\newblock {\em Neural computation}, 9(8):1735--1780, 1997.

\bibitem{BEVDet}
Junjie Huang, Guan Huang, Zheng Zhu, and Dalong Du.
\newblock {BEVDet: High-performance Multi-camera 3D Object Detection in
  Bird-Eye-View}.
\newblock {\em arXiv preprint arXiv:2112.11790}, 2021.

\bibitem{kumar2021groomed}
Abhinav Kumar, Garrick Brazil, and Xiaoming Liu.
\newblock {GrooMeD-NMS: Grouped Mathematically Differentiable NMS for Monocular
  3D Object Detection}.
\newblock In {\em Proceedings of the IEEE Conference on Computer Vision and
  Pattern Recognition}, pages 8973--8983, 2021.

\bibitem{PointPillar}
Alex~H Lang, Sourabh Vora, Holger Caesar, Lubing Zhou, Jiong Yang, and Oscar
  Beijbom.
\newblock {PointPillars: Fast Encoders for Object Detection from Point Clouds}.
\newblock In {\em Proceedings of the IEEE Conference on Computer Vision and
  Pattern Recognition}, pages 12697--12705, 2019.

\bibitem{BEVFormer}
Zhiqi Li, Wenhai Wang, Hongyang Li, Enze Xie, Chonghao Sima, Tong Lu, Qiao Yu,
  and Jifeng Dai.
\newblock {BEVFormer: Learning Bird's-Eye-View Representation from Multi-Camera
  Images via Spatiotemporal Transformers}.
\newblock {\em arXiv preprint arXiv:2203.17270}, 2022.

\bibitem{COCO}
Tsung-Yi Lin, Michael Maire, Serge Belongie, James Hays, Pietro Perona, Deva
  Ramanan, Piotr Doll{\'a}r, and C~Lawrence Zitnick.
\newblock {Microsoft COCO: Common Objects in Context}.
\newblock In {\em Proceedings of the European Conference on Computer Vision},
  pages 740--755. Springer, 2014.

\bibitem{liu2018mobile}
Mason Liu and Menglong Zhu.
\newblock {Mobile Video Object Detection with Temporally-Aware Feature Maps}.
\newblock In {\em Proceedings of the IEEE Conference on Computer Vision and
  Pattern Recognition}, pages 5686--5695, 2018.

\bibitem{liu2019looking}
Mason Liu, Menglong Zhu, Marie White, Yinxiao Li, and Dmitry Kalenichenko.
\newblock {Looking Fast and Slow: Memory-Guided Mobile Video Object Detection}.
\newblock {\em arXiv preprint arXiv:1903.10172}, 2019.

\bibitem{PETR}
Yingfei Liu, Tiancai Wang, Xiangyu Zhang, and Jian Sun.
\newblock {PETR: Position Embedding Transformation for Multi-View 3D Object
  Detection}.
\newblock {\em arXiv preprint arXiv:2203.05625}, 2022.

\bibitem{liu2021autoshape}
Zongdai Liu, Dingfu Zhou, Feixiang Lu, Jin Fang, and Liangjun Zhang.
\newblock {AutoShape: Real-Time Shape-Aware Monocular 3D Object Detection}.
\newblock In {\em Proceedings of the International Conference on Computer
  Vision}, pages 15641--15650, 2021.

\bibitem{AdamW}
Ilya Loshchilov and Frank Hutter.
\newblock {DECOUPLED WEIGHT DECAY REGULARIZATION}.
\newblock In {\em Proceedings of the International Conference on Learning
  Representations}, 2019.

\bibitem{lu2017online}
Yongyi Lu, Cewu Lu, and Chi-Keung Tang.
\newblock {Online Video Object Detection using Association LSTM}.
\newblock In {\em Proceedings of the International Conference on Computer
  Vision}, pages 2344--2352, 2017.

\bibitem{lu2021geometry}
Yan Lu, Xinzhu Ma, Lei Yang, Tianzhu Zhang, Yating Liu, Qi Chu, Junjie Yan, and
  Wanli Ouyang.
\newblock {Geometry Uncertainty Projection Network for Monocular 3D Object
  Detection}.
\newblock In {\em Proceedings of the International Conference on Computer
  Vision}, pages 3111--3121, 2021.

\bibitem{Centerfusion}
Ramin Nabati and Hairong Qi.
\newblock {CenterFusion: Center-based Radar and Camera Fusion for 3D Object
  Detection}.
\newblock In {\em Proceedings of the IEEE/CVF Winter Conference on Applications
  of Computer Vision}, pages 1527--1536, 2021.

\bibitem{VPN}
Bowen Pan, Jiankai Sun, Ho~Yin~Tiga Leung, Alex Andonian, and Bolei Zhou.
\newblock {Cross-View Semantic Segmentation for Sensing Surroundings}.
\newblock {\em IEEE Robotics and Automation Letters}, 5(3):4867--4873, 2020.

\bibitem{DD3D}
Dennis Park, Rares Ambrus, Vitor Guizilini, Jie Li, and Adrien Gaidon.
\newblock {Is Pseudo-Lidar needed for Monocular 3D Object detection?}
\newblock In {\em Proceedings of the International Conference on Computer
  Vision}, pages 3142--3152, 2021.

\bibitem{LSS}
Jonah Philion and Sanja Fidler.
\newblock {Lift, Splat, Shoot: Encoding Images from Arbitrary Camera Rigs by
  Implicitly Unprojecting to 3D}.
\newblock In {\em Proceedings of the European Conference on Computer Vision},
  pages 194--210. Springer, 2020.

\bibitem{reading2021categorical}
Cody Reading, Ali Harakeh, Julia Chae, and Steven~L Waslander.
\newblock {Categorical Depth Distribution Network for Monocular 3D Object
  Detection}.
\newblock In {\em Proceedings of the IEEE Conference on Computer Vision and
  Pattern Recognition}, pages 8555--8564, 2021.

\bibitem{PON}
Thomas Roddick and Roberto Cipolla.
\newblock {Predicting Semantic Map Representations from Images using Pyramid
  Occupancy Networks}.
\newblock In {\em Proceedings of the IEEE Conference on Computer Vision and
  Pattern Recognition}, pages 11138--11147, 2020.

\bibitem{ImageNetVID}
Olga Russakovsky, Jia Deng, Hao Su, Jonathan Krause, Sanjeev Satheesh, Sean Ma,
  Zhiheng Huang, Andrej Karpathy, Aditya Khosla, Michael Bernstein, et~al.
\newblock {ImageNet Large Scale Visual Recognition Challenge}.
\newblock {\em International Journal of Computer Vision}, 115(3):211--252,
  2015.

\bibitem{MonoDIS}
Andrea Simonelli, Samuel~Rota Bulo, Lorenzo Porzi, Manuel L{\'o}pez-Antequera,
  and Peter Kontschieder.
\newblock {Disentangling Monocular 3D Object Detection}.
\newblock In {\em Proceedings of the International Conference on Computer
  Vision}, pages 1991--1999, 2019.

\bibitem{Waymo}
Pei Sun, Henrik Kretzschmar, Xerxes Dotiwalla, Aurelien Chouard, Vijaysai
  Patnaik, Paul Tsui, James Guo, Yin Zhou, Yuning Chai, Benjamin Caine, et~al.
\newblock {Scalability in Perception for Autonomous Driving: Waymo Open
  Dataset}.
\newblock In {\em Proceedings of the IEEE Conference on Computer Vision and
  Pattern Recognition}, pages 2446--2454, 2020.

\bibitem{FCOS}
Zhi Tian, Chunhua Shen, Hao Chen, and Tong He.
\newblock {FCOS: Fully Convolutional One-Stage Object Detection}.
\newblock In {\em Proceedings of the International Conference on Computer
  Vision}, pages 9627--9636, 2019.

\bibitem{Transformer}
Ashish Vaswani, Noam Shazeer, Niki Parmar, Jakob Uszkoreit, Llion Jones,
  Aidan~N Gomez, {\L}ukasz Kaiser, and Illia Polosukhin.
\newblock {Attention is All you Need}.
\newblock {\em Advances in Neural Information Processing Systems}, 30, 2017.

\bibitem{wang2021depth}
Li Wang, Liang Du, Xiaoqing Ye, Yanwei Fu, Guodong Guo, Xiangyang Xue, Jianfeng
  Feng, and Li Zhang.
\newblock {Depth-conditioned Dynamic Message Propagation for Monocular 3D
  Object Detection}.
\newblock In {\em Proceedings of the IEEE Conference on Computer Vision and
  Pattern Recognition}, pages 454--463, 2021.

\bibitem{wang2021progressive}
Li Wang, Li Zhang, Yi Zhu, Zhi Zhang, Tong He, Mu Li, and Xiangyang Xue.
\newblock {Progressive Coordinate Transforms for Monocular 3D Object
  Detection}.
\newblock In {\em Advances in Neural Information Processing Systems}, 2021.

\bibitem{FCOS3D}
Tai Wang, Xinge Zhu, Jiangmiao Pang, and Dahua Lin.
\newblock {FCOS3D: Fully Convolutional One-Stage Monocular 3D Object
  Detection}.
\newblock {\em arXiv preprint arXiv:2104.10956}, 2021.

\bibitem{PGD}
Tai Wang, Xinge Zhu, Jiangmiao Pang, and Dahua Lin.
\newblock {Probabilistic and Geometric Depth: Detecting Objects in
  Perspective}.
\newblock {\em arXiv preprint arXiv:2107.14160}, 2021.

\bibitem{DETR3D}
Yue Wang, Vitor Guizilini, Tianyuan Zhang, Yilun Wang, Hang Zhao, and Justin
  Solomon.
\newblock {DETR3D: 3D Object Detection from Multi-view Images via 3D-to-2D
  Queries}.
\newblock {\em arXiv preprint arXiv:2110.06922}, 2021.

\bibitem{Second}
Yan Yan, Yuxing Mao, and Bo Li.
\newblock {SECOND: Sparsely Embedded Convolutional Detection}.
\newblock {\em Sensors}, 18(10):3337, 2018.

\bibitem{PYVA}
Weixiang Yang, Qi Li, Wenxi Liu, Yuanlong Yu, Yuexin Ma, Shengfeng He, and Jia
  Pan.
\newblock {Projecting Your View Attentively: Monocular Road Scene Layout
  Estimation via Cross-View Transformation}.
\newblock In {\em Proceedings of the IEEE Conference on Computer Vision and
  Pattern Recognition}, pages 15536--15545, 2021.

\bibitem{TAFVOD}
Wenfei Yang, Bin Liu, Weihai Li, and Nenghai Yu.
\newblock {Tracking Assisted Faster Video Object Detection}.
\newblock In {\em 2019 IEEE International Conference on Multimedia and Expo
  (ICME)}, pages 1750--1755, 2019.

\bibitem{CenterPoint3D}
Tianwei Yin, Xingyi Zhou, and Philipp Krahenbuhl.
\newblock {Center-based 3D Object Detection and Tracking}.
\newblock In {\em Proceedings of the IEEE Conference on Computer Vision and
  Pattern Recognition}, pages 11784--11793, 2021.

\bibitem{zhang2021objects}
Yunpeng Zhang, Jiwen Lu, and Jie Zhou.
\newblock {Objects are Different: Flexible Monocular 3D Object Detection}.
\newblock In {\em Proceedings of the IEEE Conference on Computer Vision and
  Pattern Recognition}, pages 3289--3298, 2021.

\bibitem{CenterNet}
Xingyi Zhou, Dequan Wang, and Philipp Kr{\"a}henb{\"u}hl.
\newblock {Objects as Points}.
\newblock {\em arXiv preprint arXiv:1904.07850}, 2019.

\bibitem{zhou2021monocular}
Yunsong Zhou, Yuan He, Hongzi Zhu, Cheng Wang, Hongyang Li, and Qinhong Jiang.
\newblock {Monocular 3D Object Detection: An Extrinsic Parameter Free
  Approach}.
\newblock In {\em Proceedings of the IEEE Conference on Computer Vision and
  Pattern Recognition}, pages 7556--7566, 2021.

\bibitem{CBGS}
Benjin Zhu, Zhengkai Jiang, Xiangxin Zhou, Zeming Li, and Gang Yu.
\newblock {Class-balanced Grouping and Sampling for Point Cloud 3D Object
  Detection}.
\newblock {\em arXiv preprint arXiv:1908.09492}, 2019.

\bibitem{zhu2018towards}
Xizhou Zhu, Jifeng Dai, Lu Yuan, and Yichen Wei.
\newblock {Towards High Performance Video Object Detection}.
\newblock In {\em Proceedings of the IEEE Conference on Computer Vision and
  Pattern Recognition}, pages 7210--7218, 2018.

\bibitem{zhu2017flow}
Xizhou Zhu, Yujie Wang, Jifeng Dai, Lu Yuan, and Yichen Wei.
\newblock {Flow-Guided Feature Aggregation for Video Object Detection}.
\newblock In {\em Proceedings of the International Conference on Computer
  Vision}, pages 408--417, 2017.

\bibitem{zou2021devil}
Zhikang Zou, Xiaoqing Ye, Liang Du, Xianhui Cheng, Xiao Tan, Li Zhang, Jianfeng
  Feng, Xiangyang Xue, and Errui Ding.
\newblock {The Devil Is in the Task: Exploiting Reciprocal
  Appearance-Localization Features for Monocular 3D Object Detection}.
\newblock In {\em Proceedings of the International Conference on Computer
  Vision}, pages 2713--2722, 2021.

\end{thebibliography}
}
\end{document}